\documentclass[10pt,twocolumn,letterpaper]{article}

\usepackage{cvpr}              

\usepackage{graphicx}
\usepackage{amsmath}
\usepackage{amssymb}
\usepackage{booktabs}
\usepackage{multirow}
\usepackage{mathrsfs}
\usepackage{subcaption}
\usepackage{mathtools}
\usepackage{bm}
\usepackage{pifont}
\usepackage{xcolor}
\usepackage{wrapfig}

\usepackage[normalem]{ulem}

\DeclareMathOperator\RR{\mathbb{R}}
\DeclarePairedDelimiter\p()
\DeclarePairedDelimiter\s\{\}
\newcommand{\YES}{\ding{52}}

\newcommand{\DFFT}[1]{DFFT\textsubscript{#1}}

\newcommand{\BackboneFunction}{F}
\newcommand{\BackbonePreprocessFunction}{\BackboneFunction_\mathrm{embed}}
\newcommand{\BackboneBlockFunction}{\BackboneFunction_\mathrm{block}}
\newcommand{\BackboneStageFunction}{\BackboneFunction_\mathrm{stage}}
\newcommand{\BackboneSemanticAttentionFunction}{\BackboneFunction_\mathrm{se\textrm{-}att}}
\newcommand{\ScaleEncoderFunction}{S}
\newcommand{\ScaleEncoderBlockFunction}{\ScaleEncoderFunction_\mathrm{att}}
\newcommand{\TaskEncoderFunction}{T}
\newcommand{\TaskEncoderGroupAttentionFunction}{\TaskEncoderFunction_\mathrm{group}}
\newcommand{\TaskEncoderGlobalAttentionFunction}{\TaskEncoderFunction_\mathrm{global}}
\newcommand{\UpsampleFunction}{\mathrm{up}}
\newcommand{\DownsampleFunction}{\mathrm{down}}

\newcommand{\BackboneFeature}{\bm{f}^\mathrm{dot}}
\newcommand{\BackboneBlockFeature}{\hat{\bm{f}}}
\newcommand{\BackboneSemanticAttentionFeature}{\widetilde{\bm{f}}}
\newcommand{\ScaleEncoderFeature}{\bm{s}^\mathrm{sae}}
\newcommand{\TaskEncoderFeature}{\bm{t}}
\newcommand{\ClassificationFeature}{\TaskEncoderFeature^\mathrm{cls}}
\newcommand{\RegressionFeature}{\TaskEncoderFeature^\mathrm{reg}}

\usepackage[capitalize]{cleveref}
\crefname{section}{Sec.}{Secs.}
\Crefname{section}{Section}{Sections}
\Crefname{table}{Table}{Tables}
\crefname{table}{Tab.}{Tabs.}

\begin{document}
\newcommand{\etals} {\textit{et al.}}

\title{
Efficient Decoder-free  Object Detection with Transformers}

\author{Peixian Chen$^\dag$\footnotemark[1], ~~
Mengdan Zhang$^\dag$\footnotemark[1], ~~
Yunhang Shen$^\dag$, ~~ 
Kekai Sheng$^\dag$,~~ 
Yuting Gao$^\dag$, ~~ \\
Xing Sun$^\dag$, ~~ 
Ke Li$^\dag$, ~~
Chunhua Shen$^\ddag$ \\
[0.2cm]
$^\dag$ Tencent Youtu Lab  ~~~ 
$^\ddag $ Zhejiang University
}
\maketitle

\begin{abstract}
Vision transformers (ViTs) are changing the landscape of object detection approaches. A natural usage of ViTs in detection is to replace the CNN-based backbone with a transformer-based backbone, which is straightforward and effective, with the price of bringing considerable computation burden for inference. More subtle usage is the DETR family, which eliminates the need for many hand-designed components in object detection but introduces a decoder demanding an extra-long time to converge. As a result, transformer-based object detection can not prevail in large-scale applications. To overcome these issues, we propose a novel decoder-free fully transformer-based (DFFT) object detector, achieving high efficiency in both training and inference stages, for the first time. We simplify objection detection into an encoder-only single-level anchor-based dense prediction problem by centering around two entry points: 1) Eliminate the training-inefficient decoder and leverage two strong encoders to preserve the accuracy of single-level feature map prediction; 2) Explore low-level semantic features for the detection task with limited computational resources. In particular, we design a novel lightweight detection-oriented transformer backbone that efficiently captures low-level features with rich semantics based on a well-conceived ablation study. Extensive experiments on the MS COCO benchmark demonstrate that \DFFT{SMALL} outperforms DETR by $2.5\%$ AP with $28\%$ computation cost reduction and more than $10\times$ fewer training epochs. Compared with the cutting-edge anchor-based detector RetinaNet, \DFFT{SMALL} obtains over $5.5\%$ AP gain while cutting down $70\%$ computation cost. The code is available at \url{https://github.com/Pealing/DFFT.}
\end{abstract}

\renewcommand{\thefootnote}{\fnsymbol{footnote}}
\footnotetext[1]{These authors contributed equally to this work.}

\section{Introduction}
\begin{figure}[t]
    \includegraphics[width=.8\linewidth]{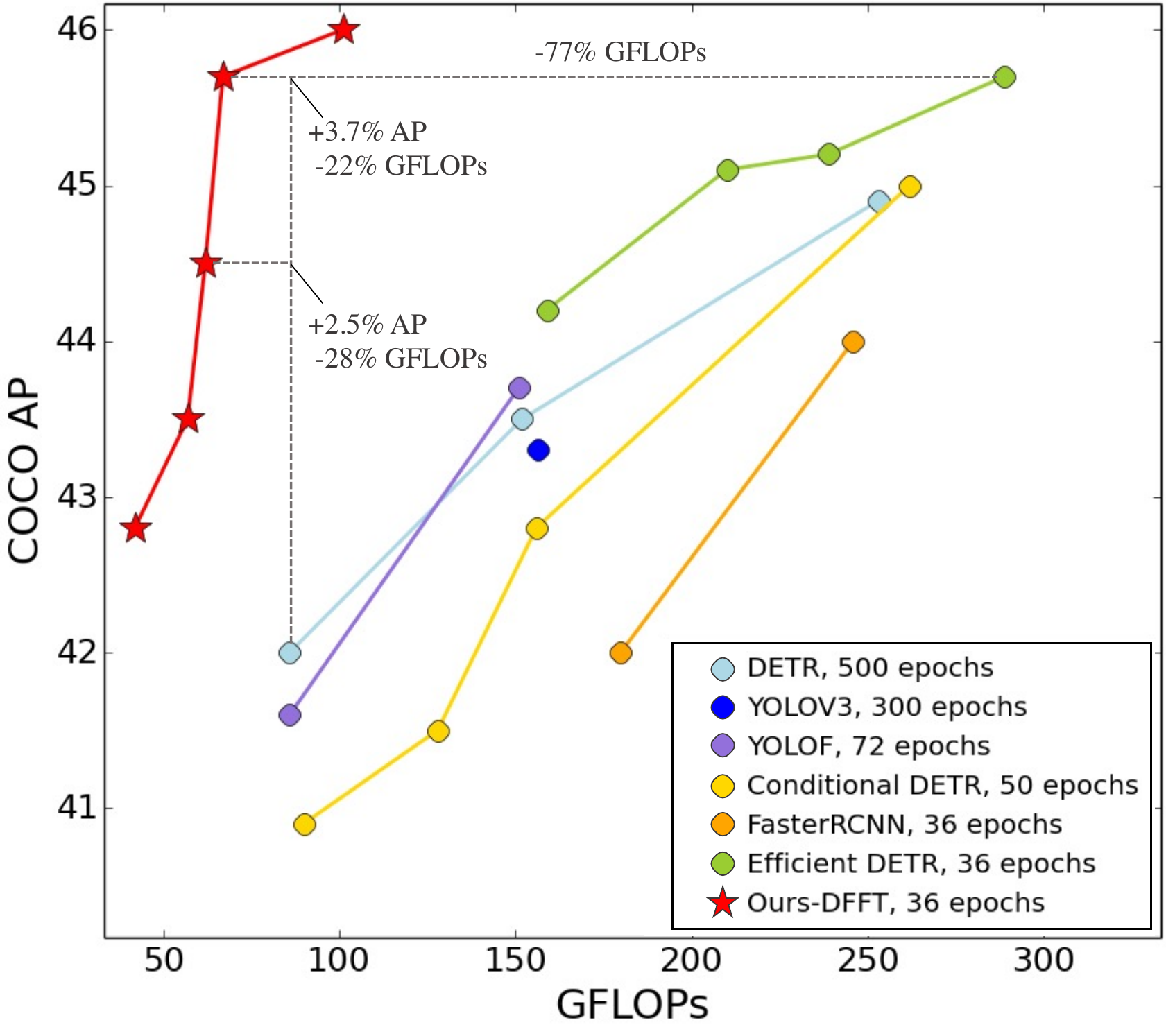}
    \caption{The trade-off between performance (AP) and efficiency (Epochs \& GFLOPs) for detection methods. With a lightweight detection-oriented backbone and a decoder-free single-level dense prediction module, \DFFT{MEDIUM} gets 22\% faster inference, more than $10\times$ fewer training epochs and 3.7\% higher AP than DETR, cuts down the 77\% GFLOPs from Efficient DETR. }
    \label{fig:fig1}
\end{figure}

Object detection is a classic computer vision task aiming to locate and recognize objects in natural images. Recently, vision transformers~\cite{SwinTransformer,PVT,MobileFormer,XCiT} have been widely developed as powerful backbones in traditional detection frameworks such as Faster RCNN~\cite{FasterRCNN}, Mask RCNN~\cite{MaskRCNN}, and RetinaNet~\cite{RetinaNet}. However, these transformer-based detectors achieve high precision at the expense of computational efficiency (\emph{e.g.}, at least 200 GFLOPs), which precludes their use in real-world applications with limited resources. DETR~\cite{DETR} is a pioneering work that addresses this issue by using an encoder-decoder transformer design that reduces object detection to an end-to-end set prediction problem. DETR's novel decoder helps object queries attend to diverse regions of interest on single-level representation, considerably boosting inference efficiency (ranging from 86 to 253 GFLOPs). Unfortunately, this enhancement comes at the expense of around 10$\times$ to 20$\times$ slower training convergence. As a result, it remains open whether transformer-based detectors can attain high precision without losing efficiency in training and inference stages.

Recent work in the DETR family has mainly focused on improving the delayed convergence induced by the decoder. They augment object queries in the decoder with explicit spatial priors such as reference points~\cite{DeformableDETR}, anchor points~\cite{AnchorDETR,TSP}, RPN proposals~\cite{EfficientDETR,FasterRCNN}, and conditional spatial embeddings~\cite{ConditionalDETR,SMCA}. However, introducing spatial priors to the decoder stage sacrifices the detector's inference efficiency, consuming more than 1.5$\times$ GFLOPs. It also raises the question whether the above efficient yet accurate transformer-based detector inevitably needs a decoder.

In this paper, we build a novel detection architecture named DFFT: \textbf{D}ecoder-\textbf{F}ree \textbf{F}ully \textbf{T}ransformer-based object detector, which achieves both higher accuracy and better training-inference efficiency across a spectrum of low resource constraints (\emph{e.g.}, from $40$ to $100$ GFLOPs) as shown in \Cref{fig:fig1}. Based on a well-conceived analysis of how different transformer architectures (\emph{e.g.}, attention components' type, position and linkage) involved in the backbone, feature fusion, and class/box network, impact the trade-off between detection performance and efficiency, DFFT simplifies the whole object detection pipeline to an encoder-only single-level anchor-based dense prediction task. Specifically, our design of DFFT centers around two entry points:

\emph{Entry Point 1: Eliminate the training-inefficient decoder and leverage two strong encoders to preserve the accuracy of single-level feature map prediction.}
To design a light-weight detection pipeline comparable to DETR and maintain high training efficiency, we eliminate the training-inefficient decoder and propose two strong transformer encoders in the feature fusion and class/box network to avoid performance decline after trimming the decoder. Benefiting from two strong encoders, DFFT conducts anchor-based dense prediction only on a single-level feature map, ensuring training and inference efficiency while maintaining high accuracy. (1) The \emph{scale-aggregated encoder} summarizes multi-scale cues to one feature map by progressively analyzing global spatial and semantic relations of two consecutive feature maps. Thus, instances of various scales are easily detected on the single feature map, avoiding exhaustive search across network layers. (2) The \emph{task-aligned encoder} enables DFFT to conduct  classification and regression simultaneously in a coupled head. By taking advantage of group channel-wise attention, it resolves the learning conflicts from the two tasks and provides consistent predictions~\cite{SiblingHead,RevisitingRCNN}.  
\emph{Entry Point 2: Explore low-level semantic features as much as possible for the detection task with limited computational resources.}
We design a strong and efficient detection-oriented transformer backbone after an in-depth study on different characteristics of transformer attention components (e.g., spatial-wise attention and channel-wise
attention).  Furthermore, we propose to incorporate semantic-augmented attention modules into several stages of the backbone to capture rich low-level semantics. Low-level semantics from different stages help the detector distinguish distractors in detail. Such design is quite different from common backbones~\cite{SwinTransformer,PVT,XCiT}  that are dedicated to learn final high-level semantics for the classification task. 


Finally, we conduct comprehensive experiments to verify the superiority of DFFT as well as the effectiveness of all the above designs.  Compared to the foundation work deformable DETR~\cite{DeformableDETR}, DFFT achieves 61\% inference acceleration, 28\% training acceleration, and 1.9\% AP gain.

\section{Related Work}
\subsection{One-stage and Two-stage Detectors}
Mainstream detectors exploit two types of anchors: anchor boxes~\cite{FasterRCNN,RetinaNet} and anchor points~\cite{FCOS,CenterNet}. Anchors are generated at the center of each sliding-window position to offer candidates for objects. Typical one-stage detectors~\cite{YOLOF,CenterNet,FCOS,YOLOX,RetinaNet} directly predict categories and offsets of anchors for the whole feature maps, while two-stage detectors~\cite{MaskRCNN,FasterRCNN,CascadeRCNN} first generate region proposals from dense anchor boxes by a Region Proposal Network~(RPN)~\cite{FasterRCNN}
and then refine the detection for each proposed region afterward.

One main challenge in object detection is to represent objects at vastly different scales effectively. Both one-stage and two-stage detectors overcome it with multi-scale features and multi-level predictions. 
FPN~\cite{FPN} is widely used in these detectors~\cite{FCOS,MaskRCNN,RetinaNet}, which builds feature pyramid by sequentially combining two adjacent layers in feature hierarchy in backbone model with top-down and lateral connections. Later CNN-based designs on cross-scale connections use bottom-up paths~\cite{PANet}, U-shape modules~\cite{M2Det}, and the neural architecture search~\cite{NASFPN,AutoFPN}.
YOLOF~\cite{YOLOF} provides an alternative solution, which exploits dilated encoder to detect all objects on single-level features.
In contrast, our DFFT introduces large receptive fields to cover large objects based on the transformer's global relation modeling, and meanwhile, aggregates low-level semantics through the scale-aggregated encoder. Such designs enable DFFT to achieve superior detection performance.

Recently, transformer-based backbones~\cite{MobileFormer,PVT,SwinTransformer,XCiT} have shown superior performance in object detection based on standard frameworks such as MaskRCNN~\cite{MaskRCNN,CascadeRCNN} and RetinaNet~\cite{RetinaNet}.
However, these backbones are usually directly plugged into the framework without regard for the effects of replacing CNN with transformers. These methods typically consume enormous computation costs (e.g., over 300 GFLOPs for MobileFormer~\cite{MobileFormer}). DFFT is the first method to explore efficient and fully transformer-based  detection.

\begin{figure*}[tbp]
\centering
\begin{subfigure}[t]{0.31\linewidth}
	\centering
	\includegraphics[width=1.\linewidth]{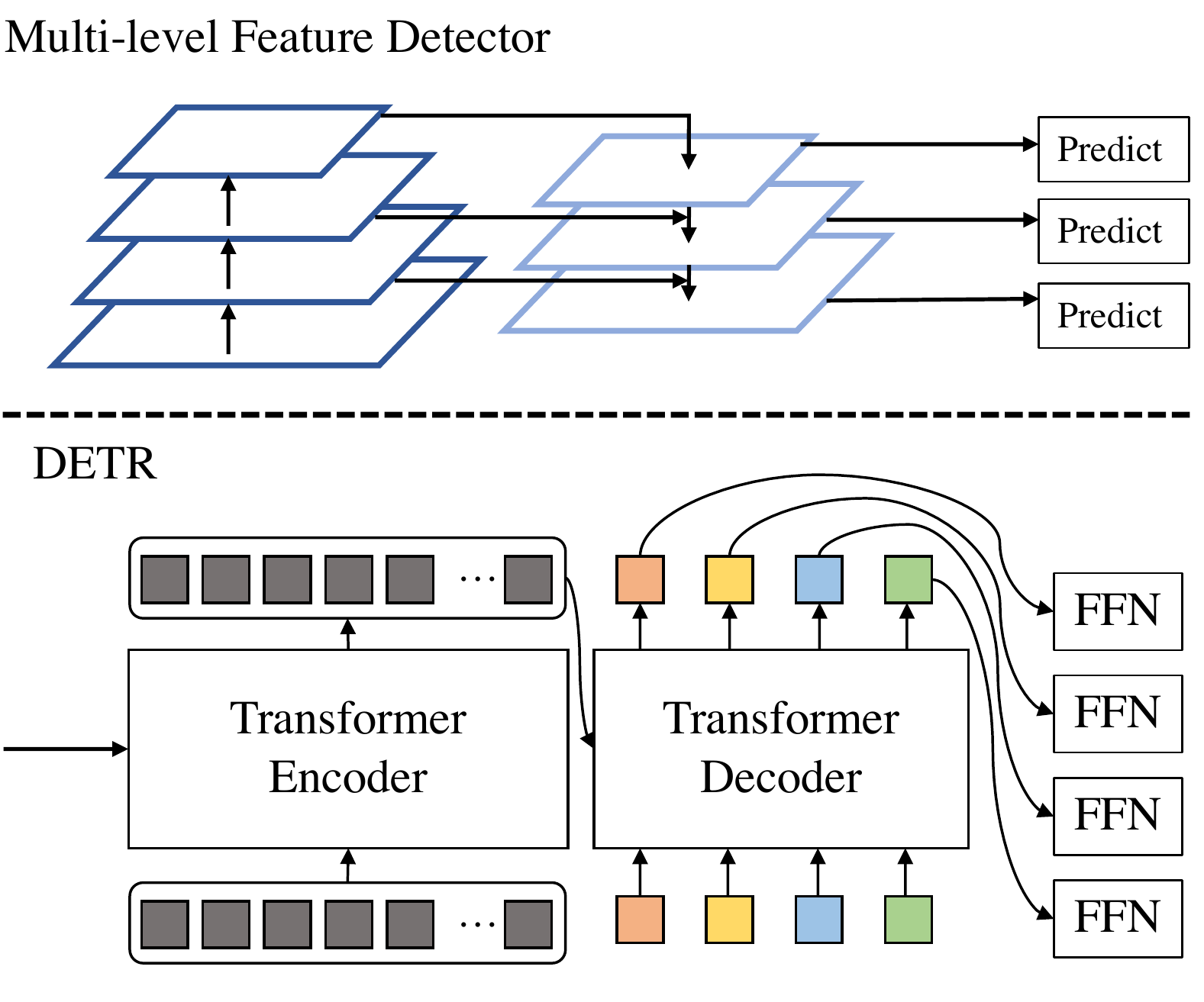}
	\caption{Overview of existing detection methods.}
	\label{fig:othersframework}
\end{subfigure}
\hfill
\begin{subfigure}[t]{0.63\linewidth}
	\centering
	\includegraphics[width=1.\linewidth]{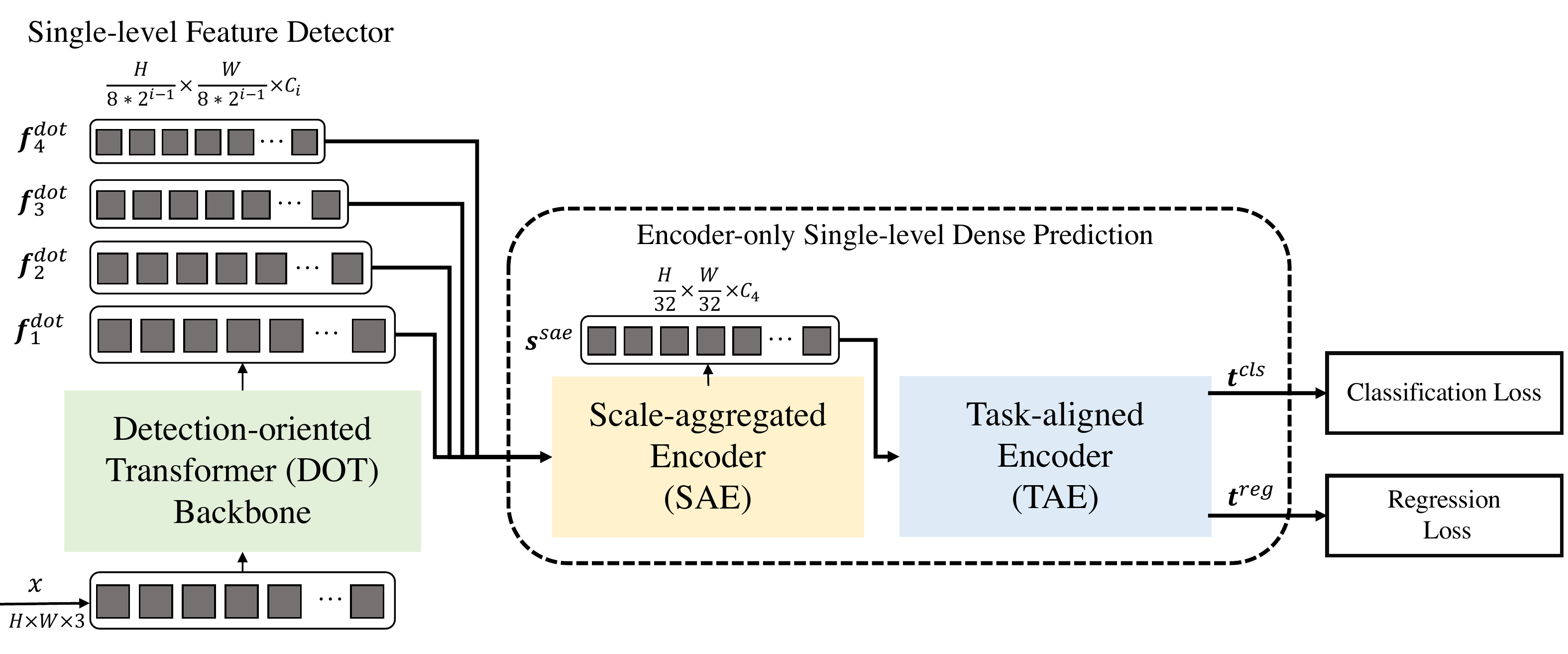}
	\caption{Overview of our proposed DFFT.}
	\label{fig:framework}
\end{subfigure}
\caption{Overview of existing detection methods and our proposed Decoder-Free Fully Transformer-based (DFFT) detector. \textbf{(a)} Existing methods either rely on multi-level feature detectors~\cite{RetinaNet} or adopt the DETR framework~\cite{DETR}. \textbf{(b)} Our DFFT simplifies object detection to an encoder-only single-level dense prediction framework. 
The proposed two strong encoders enables us to conduct fast but accurate inference on a single-level feature map, outperforming existing multi-level feature detectors. The design also  trims the training-inefficient decoder for more than $10 \times$ training acceleration over DETR.
We also propose a novel lightweight detection-oriented transformer backbone to capture richer low-level semantic features and further boost object detection.}
\label{fig:ALL}
\vspace{-1em}
\end{figure*}

\subsection{End-to-end Detectors}
End-to-end detectors~\cite{DETR,DeformableDETR,YOLOS,ConditionalDETR,SMCA} remove the complicated post-processing like NMS and achieve one-to-one matching between the target and the candidate by the Hungarian algorithm. DETR~\cite{DETR} uses an encoder-decoder transformer framework. The transformer encoder processes the flattened deep features from the CNN backbone. The non-autoregressive decoder takes the encoder’s outputs and a set of learned object query vectors as the input, and predicts the category labels and bounding boxes accordingly. The decoder's cross-attention module attend to different locations in the image for different object queries, which requires high-quality content embeddings and thus training costs. DETR needs a long training process ($500$ epochs) and is not suitable for small objects. Deformable DETR~\cite{DeformableDETR} accelerates the convergence via learnable sparse sampling and multi-scale deformable encoders. It generates a reference point for each object query and uses deformable attention to make each reference point only focus on a small fixed set of sampling points. Anchor DETR~\cite{AnchorDETR} exploits anchor points to accelerate training. Conventional designs such as RPN~\cite{EfficientDETR}, RCNN, and FCOS~\cite{TSP} are also used to optimize the DETR framework. Although better performance and fast convergence are achieved, the computation cost increases significantly (e.g., $2\times$ GFLOPs in deformable DETR) due to multi-scale feature encoding~\cite{DeformableDETR,TSP}. Moreover, dense priors such as reference points~\cite{DeformableDETR}, anchor points~\cite{AnchorDETR,TSP}, proposals~\cite{TSP}, and conditional spatial embedding~\cite{ConditionalDETR} are introduced to optimize the DETR pipeline for fast convergence. It shows that DETR is not the only solution for efficient, fully transformer-based detectors.
Compared with them, DFFT trims the decoder, also ensures both fast training convergence and inference while maintains comparable performance.

\begin{figure*}[tbp]
\centering
\begin{subfigure}[t]{0.44\linewidth}
	\centering
	\includegraphics[width=1.02\linewidth]{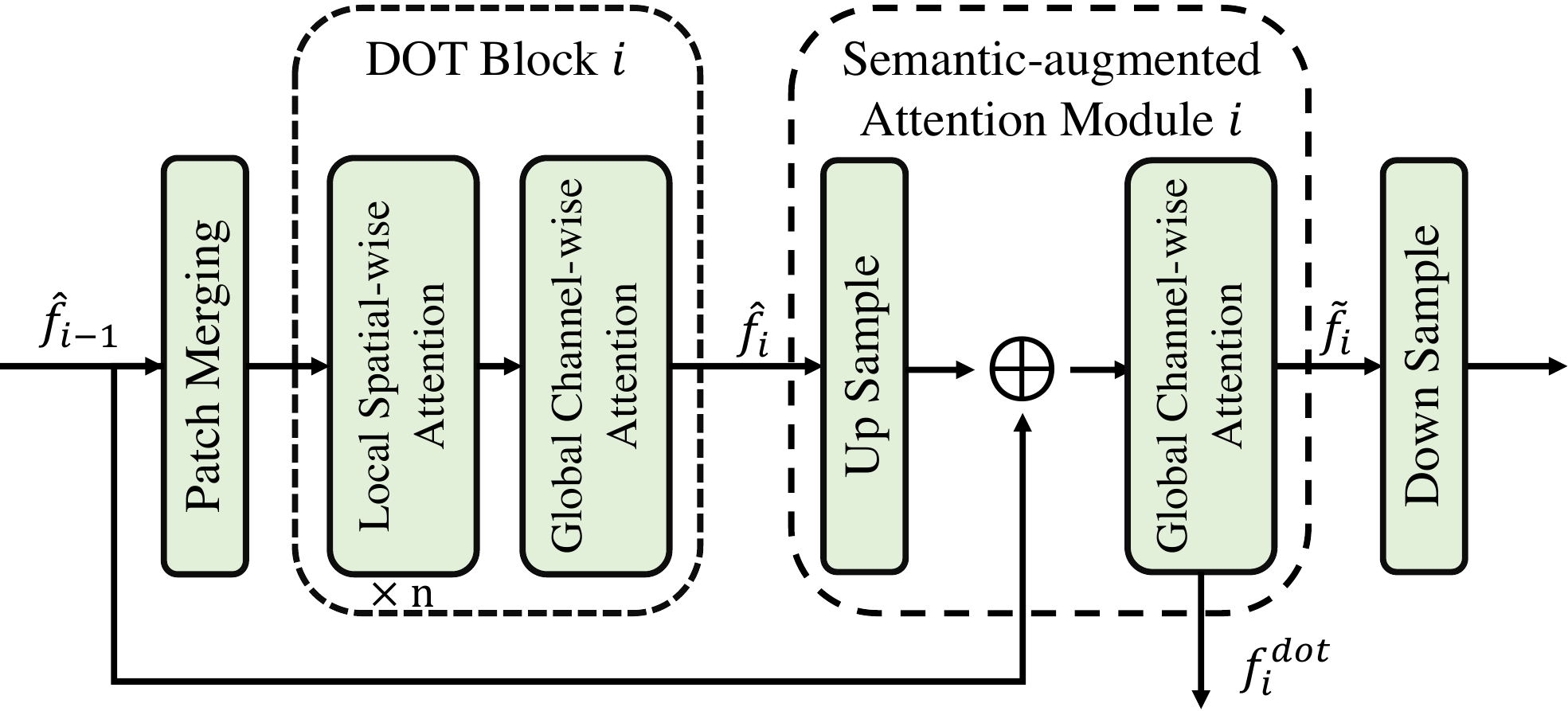}
	\caption{\centering The $i$-th DOT Backbone Stage}
	\label{fig:DOT}
\end{subfigure}
\hfill
\begin{subfigure}[t]{0.23\linewidth}
	\centering
	\includegraphics[width=.98\linewidth]{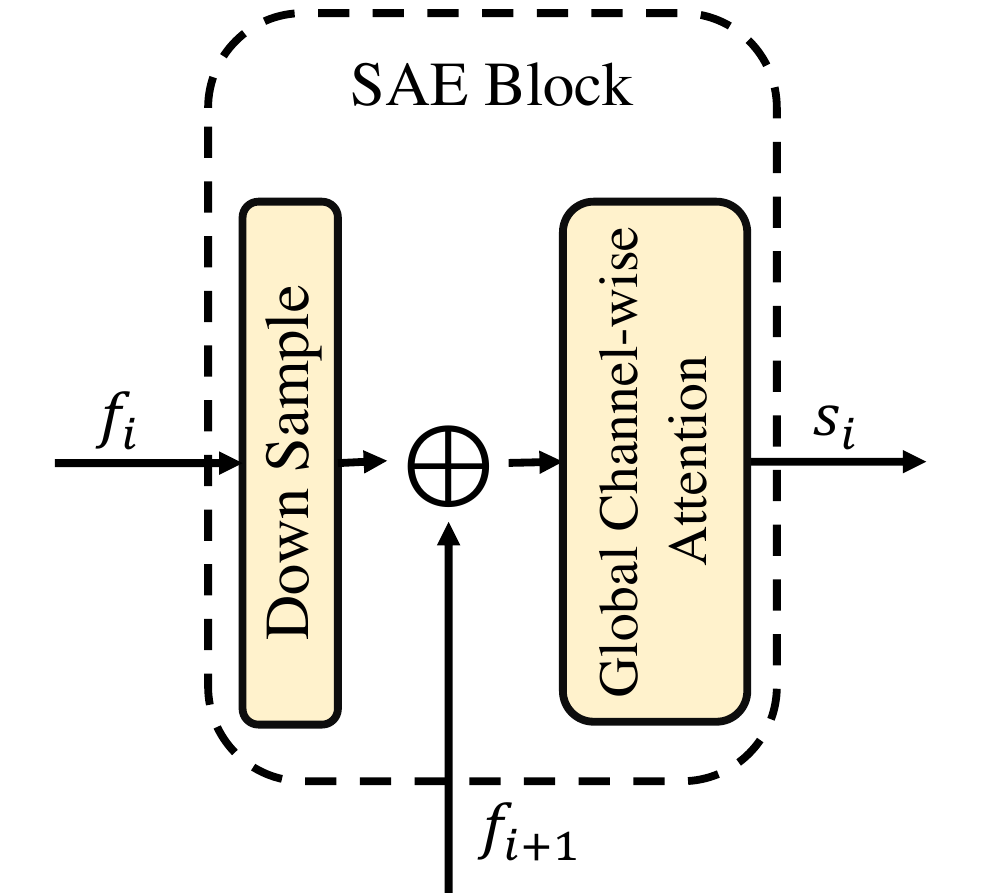}
	\caption{\centering SAE}
	\label{fig:SAE}
\end{subfigure}
\hfill
\begin{subfigure}[t]{0.3\linewidth}
	\centering
	\includegraphics[width=1.\linewidth]{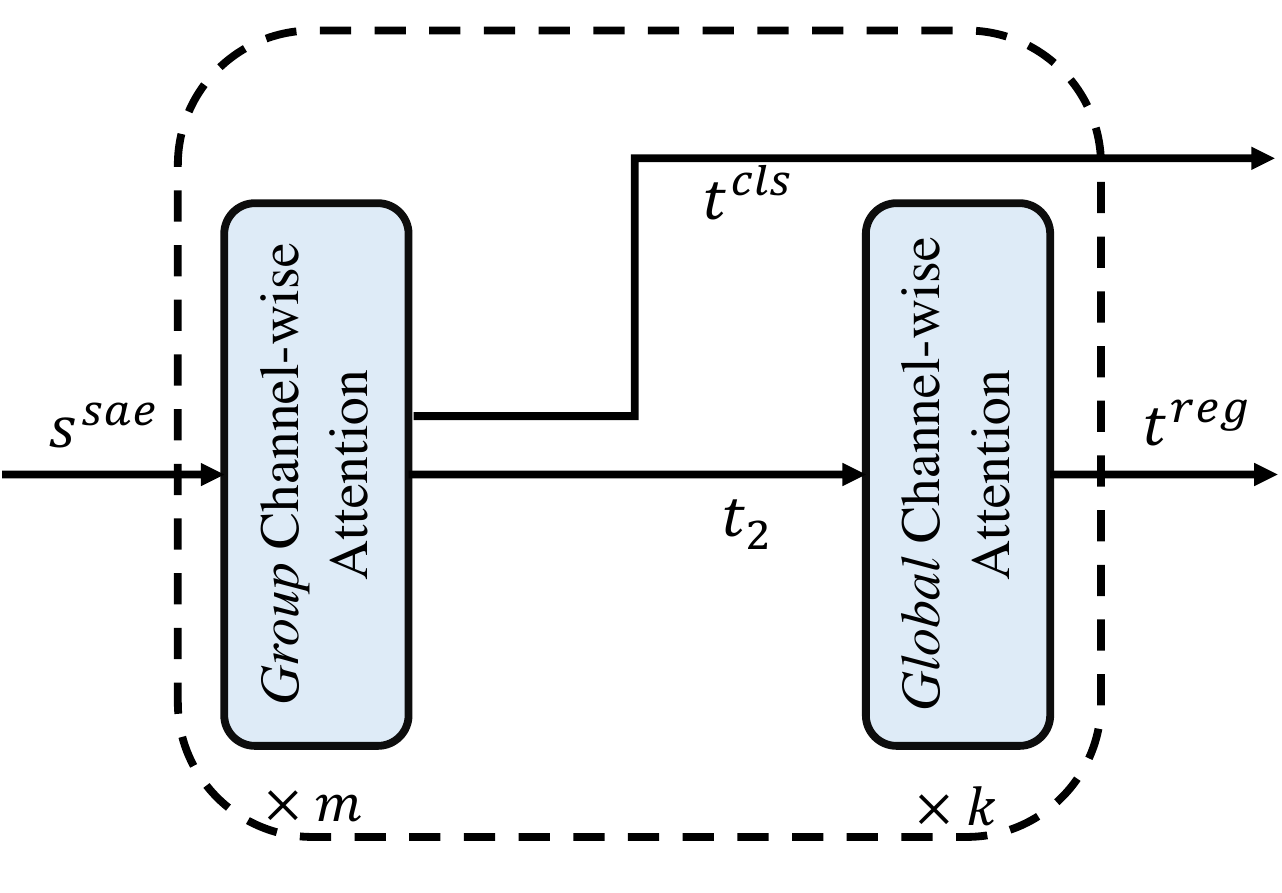}
	\caption{\centering TAE}
	\label{fig:TAE}
\end{subfigure}
\caption{Illustration of the three major modules in our proposed DFFT. DFFT contains a light-weight \textbf{D}etection-\textbf{O}riented \textbf{T}ransformer backbone with four DOT stages to extract features with rich semantic information, a \textbf{S}cale-\textbf{A}ggregated \textbf{E}ncoder (SAE) with three SAE blocks to aggregate multi-scale features into one feature map for efficiency, and a \textbf{T}ask-\textbf{A}ligned \textbf{E}ncoder (TAE) to resolve conflicts between classification and regression tasks in the coupled detection head.}
\vspace{-1.5em}
\end{figure*}

\section{Method}
\label{sec:method}

In this section, we introduce DFFT, an efficient Decoder-Free Fully Transformer-based object detector. An overview of DFFT is illustrated in \Cref{fig:framework}.
The Detection-oriented Transformer backbone $\BackboneFunction$ extracts features at four scales and sends them to the following encoder-only single-level dense prediction module. The prediction module first aggregates the multi-scale feature into a single feature map through the Scale-Aggregated Encoder $\ScaleEncoderFunction$. Then we use the Task-Aligned Encoder $\TaskEncoderFunction$ to align the feature for classification and regression tasks simultaneously for higher inference efficiency.


\subsection{Detection-oriented Transformer Backbone}

Detection-oriented transformer (DOT) backbone aims to extract multi-scale features with strong semantics. 
As shown in \Cref{fig:DOT}, it hierarchically stacks one embedding module and four DOT stages, where a novel semantic-augmented attention module aggregates the low-level semantic information of every two consecutive DOT stages.
For each input image $\bm{x} \in \mathbb{R}^{H \times W \times 3}$, the DOT backbone extracts features at four different scales:
\begin{equation}
    \BackboneFeature_1, \BackboneFeature_2, \BackboneFeature_3, \BackboneFeature_4 = \BackboneFunction(\bm{x}),
\end{equation}
where $\BackboneFeature_i\in\RR^{\frac{H}{8\cdot 2^{i-1}} \times \frac{W}{8\cdot 2^{i-1}} \times C_i}$ is the $i$-th feature with $C_i$ channels for $i\in\s{1,2,3,4}$. In what follows, we expand the formalization of the DOT backbone function $\BackboneFunction$.

\noindent\textbf{Embedding Module.}
For an input image $\bm{x} \in \mathbb{R}^{H \times W \times 3}$, we first  divide it into $\frac{H\times W}{8\times8}$ patches and feed these patches to a linear projection to obtain patch embeddings $\BackboneBlockFeature_0$ of size $\frac{H}{8} \times \frac{W}{8} \times C_1$, written as
\begin{equation}
\label{eq:embedding}
    \BackboneBlockFeature_0
    = 
    \BackbonePreprocessFunction(\bm{x})
    \in
    \RR^{\frac{H}{8} \times \frac{W}{8} \times C_1},
\end{equation}
where $\BackbonePreprocessFunction$ is the embedding module described above.

\noindent\textbf{DOT Block.}
Each DOT stage contains one DOT block $\BackboneBlockFunction$, designed to efficiently capture both the local spatial and the global semantic relations at each scale. When processing high-resolution feature maps in dense prediction, conventional transformer blocks reduce computational costs by replacing the multi-head self-attention (MSA) layer with the local spatial-wise attention layer, such as spatial-reduction attention (SRA)~\cite{PVT} and shifted window-based multi-head self-attention (SW-MSA)~\cite{SwinTransformer}. However, this design sacrifices detection performance as it only extracts multi-scale features with limited low-level semantics.

To mitigate this shortcoming, our DOT block includes multiple SW-MSA blocks~\cite{SwinTransformer} and one global channel-wise attention block~\cite{XCiT}, as illustrated in the first part of \Cref{fig:DOT}. Note that each attention block contains an attention layer and an FFN layer, and we omit the FFN layer in each attention block of \Cref{fig:ALL} to simplify the illustration. We denote by $\BackboneBlockFeature_i$ the DOT block's output feature at the $i$-th DOT stage. We find that placing a lightweight channel-wise attention layer behind consecutive local spatial-wise attention layers can benefit deducing object semantics at each scale.

\noindent\textbf{Semantic-Augmented Attention.}
While the DOT block has enhanced the semantic information in low-level features through the global channel-wise attention, semantics can be improved even further to benefit the detection task. Thus, we propose a novel \emph{semantic-augmented attention} (SAA) module $\BackboneSemanticAttentionFunction$, which 
exchanges semantic information between two consecutive scale levels and augments their features. SAA consists of an up-sampling layer and a global channel-wise attention block.
We incorporate SAA into every two consecutive DOT blocks, as illustrated in the second part of \Cref{fig:DOT}. 
Formally, SAA takes the outputs from the current DOT block and the former DOT stage, and then returns the semantic augmented feature, which is sent to the next DOT stage and also contributes to the final multi-scale feature $\BackboneFeature_i$.  We denote by $\BackboneSemanticAttentionFeature_i$ SAA's output feature at the $i$-th DOT stage.

\noindent\textbf{DOT Stage.}
The final DOT backbone contains four DOT stages $\BackboneStageFunction$ where each stage consists of one DOT block and one SAA module (except the first stage).
Specifically, the first stage contains one DOT block and no SAA module, because the inputs of the SAA module are from two consecutive DOT stages. Each of the remaining three stages contains a patch merging module to reduce the number of patches similar to \cite{SwinTransformer}, a DOT block, and a SAA module followed by a down-sampling layer to recover the input dimension, as shown in \Cref{fig:DOT}.
Thus, the formulation of the DOT block in the $i$-th stage  can be defined as
\begin{equation}
    \BackboneBlockFeature_{i} = \begin{cases}
        \BackboneBlockFunction(\BackboneBlockFeature_{i-1}), & i = 1, 2 \\
        \BackboneBlockFunction(\DownsampleFunction(\BackboneSemanticAttentionFeature_{i-1})), & i = 3, 4
    \end{cases}
\end{equation}
where $\DownsampleFunction$ denotes the downsampling function.

The $i$-th stage's SAA module can be defined as
\begin{equation}
 \BackboneSemanticAttentionFeature_i = \begin{cases}
     \BackboneSemanticAttentionFunction\p[\Big]{\UpsampleFunction(\BackboneBlockFeature_i) + \BackboneBlockFeature_{i-1}}, & i = 2 \\
     \BackboneSemanticAttentionFunction\p[\Big]{\UpsampleFunction(\BackboneBlockFeature_i) +\BackboneSemanticAttentionFeature_{i-1}}, & i=3, 4
         \end{cases}
\end{equation}
where $\UpsampleFunction$ denotes the upsampling function.
 
The final multi-scale feature from the DOT backbone can be written as
\begin{equation}
    \BackboneFeature_{i} = \begin{cases}
        \BackboneSemanticAttentionFeature_{i+1}, & i = 1, 2, 3 \\
        \BackboneBlockFeature_i, & i = 4
    \end{cases}
\end{equation}

\subsection{Encoder-only Single-level Dense Prediction}
This module is designed to improve both the inference and training efficiency of the fully transformer-based object detector with two novel encoders. It first uses the \emph{scale-aggregated} encoder (SAE) to aggregate the multi-scale features $\BackboneFeature_i$ from the DOT backbone into one feature map $\ScaleEncoderFeature$. After that, it uses the \emph{task-aligned} encoder (TAE) to generate aligned classification feature $\ClassificationFeature$ and regression feature $\RegressionFeature$ simultaneously in a single head.

%
\noindent\textbf{Scale-aggregated Encoder.}
We design this encoder with three SAE blocks, as illustrated in \Cref{fig:SAE}. Each SAE block takes two features as the input and aggregates the features step by step across all SAE blocks. We set the scale of final aggregated feature to $\frac{H}{32} \times \frac{W}{32}$ to balance the detection precision and computational costs. For this purpose, the last SAE block will up-sample the input feature to $\frac{H}{32} \times \frac{W}{32}$ before aggregation. This procedure can be described as
\begin{equation}
\begin{aligned}
  \bm{s}_0 &= \BackboneFeature_1, \\
  \bm{s}_1 &= \ScaleEncoderBlockFunction \p[\big]{\DownsampleFunction(\bm{s}_0) + \BackboneFeature_2}, \\
  \bm{s}_2 &= \ScaleEncoderBlockFunction \p[\big]{\DownsampleFunction(\bm{s}_1) + \BackboneFeature_3}, \\
  \bm{s}_3 &= \ScaleEncoderBlockFunction \p[\big]{\bm{s}_2 + \UpsampleFunction(\BackboneFeature_4)},
\end{aligned}
\end{equation}
where $\ScaleEncoderBlockFunction$ is the global channel-wise attention block and $\ScaleEncoderFeature=\bm{s}_3$ is the final aggregated feature map.

\noindent\textbf{Task-aligned Encoder.}
Recent one-stage detectors~\cite{YOLOF,FCOS} perform object classification and localization independently with two separate branches (e.g., decoupled head). This two-branch design omits the interaction between two tasks and leads to inconsistent predictions~\cite{TOOD,MutualSup}.  Meanwhile, feature learning for two tasks in a coupled head usually exists conflicts~\cite{DynamicHead,SiblingHead}. We propose the task-aligned encoder which offers a better balance between learning task-interactive and task-specific features via stacking \emph{group} channel-wise attention blocks in a coupled head.

As shown in \Cref{fig:TAE}, this encoder consists of two kinds of channel-wise attention blocks. First, the stacked \emph{group} channel-wise attention blocks $\TaskEncoderGroupAttentionFunction$ align and finally split the aggregated feature $\ScaleEncoderFeature$ into two parts. Second, the \emph{global} channel-wise attention blocks $\TaskEncoderGlobalAttentionFunction$ further encode one of the two split features for the subsequent regression task. This procedure can be described as
\begin{equation}
\begin{aligned}
  \TaskEncoderFeature_1, \TaskEncoderFeature_2 &= \TaskEncoderGroupAttentionFunction(\ScaleEncoderFeature), \\
    \ClassificationFeature &= \TaskEncoderFeature_1, \\
    \RegressionFeature &= \TaskEncoderGlobalAttentionFunction(\TaskEncoderFeature_2), \\
\end{aligned}
\end{equation}
where $\TaskEncoderFeature_1,\TaskEncoderFeature_2\in\RR^{\frac{H}{32} \times \frac{W}{32} \times 256}$ are the split features, and $\ClassificationFeature\in\RR^{\frac{H}{32} \times \frac{W}{32} \times 256}$ and $\RegressionFeature\in\RR^{\frac{H}{32} \times \frac{W}{32} \times 512}$ are the final features for the classification and regression tasks, respectively. 

Specifically, the differences between the \emph{group} channel-wise attention block and the \emph{global} channel-wise attention block lie in that all the linear projections except the projections for key/query/value embeddings in the \emph{group} channel-wise attention block are conducted in two groups. Thus, features interact in attention operations while deduced separately in output projections. 

\begin{table}[tb]   
    \centering
    \footnotesize
    \resizebox{\linewidth}{!}{
    \begin{tabular}{l|cc|cc|cc}
    \toprule
    \multirow{2}{*}{Models} & \multicolumn{2}{c|}{Backbone Settings} & \multicolumn{2}{c|}{Effectiveness (\%)} & \multicolumn{2}{c}{Efficiency (GFLOPs)} \\
     & Value of $C_i$ & Number of SA & Accuracy & AP & \multicolumn{1}{c}{Backbone} & \multicolumn{1}{c}{DFFT} \\ \midrule
    \DFFT{NANO} & $\p{3, 3, 6, 9}$ & $\p{2, 2, 6, 2}$ & 80.0 &42.8 & 26 & 42 \\
     \DFFT{TINY}&$\p{4, 4, 8, 12}$ & $\p{1, 1, 5, 1}$ & 81.1 &43.5  & 39 & 57  \\
     \DFFT{SMALL}& $\p{4, 4, 8, 12}$ & $\p{2, 2, 6, 2}$ & 82.1 &44.5  & 44 & 62  \\
     \DFFT{MEDIUM}& $\p{4, 4, 7, 12}$ & $\p{2, 2, 18, 2}$ & 82.7 &45.7 & 48 & 67  \\
     \DFFT{LARGE}& $\p{6, 6, 8, 12}$ & $\p{2, 2, 18, 2}$ &83.1  &46.0  &83  &101  \\ \bottomrule
    \end{tabular}
    }
    \caption{The definition and performance of DFFT models with different magnitudes. In the backbone setting, we list the output feature's number of channels $C_i$ and the number of SA blocks in all four backbone stages. In the effectiveness evaluation, we report the accuracy of the pre-trained backbone on ImageNet and the detection AP of DFFT after training on the MS COCO dataset.}
    \label{tab:Details}
    \vspace{-1em}
\end{table}
\subsection{Miscellaneous}
Since DFFT conducts the single-level dense prediction on a single feature map, the pre-defined anchors are sparse. Applying the Max-IoU matching~\cite{RetinaNet} based on the sparse anchors will cause an imbalance problem for positive anchors, making detectors pay attention to large ground-truth boxes while ignoring the small ones when training. To overcome this problem, we use the uniform matching strategy proposed by YOLOF~\cite{YOLOF} to ensure that all ground-truth boxes uniformly match with the same number of positive anchors regardless of their sizes. Similar to the setting of most conventional detection methods~\cite{FCOS,YOLOF,RetinaNet}, our loss function consists of a focal loss for classification and a generalized IOU loss for regression.
At the inference stage, we conduct object detection efficiently based on the final aggregated feature map $\ScaleEncoderFeature$ with a single pass.
\begin{table*}[tbp]
    \centering
    \footnotesize
    \resizebox{.9\linewidth}{!}{
    \begin{tabular}{l|c|cccccc|c}    
        \toprule    
            Methods & Epochs & AP (\%) & AP\textsubscript{50} (\%) & AP\textsubscript{75} (\%) & AP\textsubscript{S} (\%) & AP\textsubscript{M} (\%) & AP\textsubscript{L} (\%) & GFLOPs \\    
        \midrule  
            Faster RCNN-FPN-R50~\cite{FasterRCNN} & 36 & 40.2 & 61.0 & 43.8 & 24.2 & 43.5 & 52.0 & 180 \\
            RetinaNet~\cite{RetinaNet} & 12 & 35.9 & 55.7 & 38.5 & 19.4 & 39.5 & 48.2 & 201 \\
            YOLOF-R50~\cite{YOLOF}  &  12 & 37.7 & 56.9 & 40.6 & 19.1 & 42.5 & 53.2 & 86 \\
            Swin-Tiny-RetinaNet~\cite{SwinTransformer} & 12 & 42.0 &- &- &- &- &- & 245 \\
            Focal-Tiny-RetinaNet~\cite{FocalTrans} & 12 & 43.7 & - & - & - & - & - & 265\\
            Mobile-Former~\cite{MobileFormer} & 12 & 34.2 & 53.4 &36.0 &19.9 &36.8 &45.3 & 322 \\
        \midrule   
            \DFFT{NANO} & 12 &39.1 &58.3 &41.7 &19.0 &42.9 &51.2 &42 \\
            \DFFT{SMALL} & 12 & 41.4 & 60.9 & 44.5 & 20.1 & 45.4 & 58.9 & 62\\
            \DFFT{MEDIUM} & 12 &42.6 &62.5 &45.5 &22.6 &46.7 &61.4 &67  \\
        \midrule   
            DETR-R50~\cite{DETR} & 500 & 42.0 & 62.4 & 44.2 & 20.5 &45.8 & 61.1 &86 \\
            WB-DETR~\cite{WBDETR} &500 &39.6 &58.4 &43.8 &18.2 &42.7 &54.9 &62 \\
            YOLOS~\cite{YOLOS}. & 150 &37.6 & - & - & - & - & -& 172 \\
            Deformable DETR~\cite{DeformableDETR} & 50 & 43.8 & 62.6 & 47.7 & 26.4 & 47.1 & 58.0 & 173 \\
            SMCA-R50~\cite{SMCA} & 50 & 43.7 & 63.6  & 47.2 & 24.2 & 47.0 & 60.4 & 152 \\ 
            Anchor DETR-DC5-R50~\cite{AnchorDETR} & 50 & 44.2 & 64.7 & 47.5 &24.7 &48.2 & 60.6 &151\\
            Conditional DETR-R50~\cite{ConditionalDETR} & 50 & 40.9 & 61.8 & 43.3 & 20.8 & 44.6 & 59.2 & 90 \\
            TSP-FCOS-R50~\cite{TSP}  & 36 & 43.1 & 62.3 & 47.0 & 26.6 & 46.8 & 55.9 & 189 \\
            Efficient DETR-R50~\cite{EfficientDETR} & 36 & 44.2 & 62.2 & 48.0 & \textbf{28.4} & 47.5 & 56.6 & 159 \\
        \midrule
            \DFFT{NANO} & 36 &42.8 &61.9 &46.2 &23.4 &46.8 &59.7 &42  \\
            \DFFT{SMALL} & 36 & 44.5 & 63.6  & 48.0 & 24.5 & 49.0 & 60.7 & 62\\
            \DFFT{MEDIUM} & 36 &\textbf{45.7} &\textbf{64.8} &\textbf{49.7} &25.5 &\textbf{50.4} &\textbf{63.1} &67  \\
        \bottomrule   
    \end{tabular}
    }
    \caption{Comparison of our DFFT and modern detection methods on the MS COCO benchmark~\cite{MSCOCO}. The table is divided into four sections from top to bottom: (1) anchor-based methods, (2) DFFT trained for 12 epochs, (3) DETR-based methods, and (4) DFFT trained for 36 epochs. 
    DFFT achieves competitive precision with significantly fewer training epochs and inference GFLOPs.}
    \label{tab:Comparison}
    \vspace{-1.1em}
\end{table*}
\section{Experiments}
We evaluate our proposed DFFT on the challenging MS COCO benchmark~\cite{MSCOCO} following the commonly used setting. It contains around 160K images of 80 categories. 
We compare DFFT with conventional one-stage/two-stage detection methods and DETR-based methods. We also provide a comprehensive ablation study to quantitatively analyze the effectiveness of each module in DFFT. The standard mean average precision (AP) metric is used to measure detection under different IoU thresholds and object scales. 

\subsection{Settings}
The DOT backbone is pre-trained on ImageNet~\cite{ImageNet} with the same setting as~\cite{SwinTransformer}.
We train DFFT with the standard $1\times$ ($12$ epochs) and $3\times$ ($36$ epochs) training configurations as introduced in~\cite{SwinTransformer}. We
use the AdamW~\cite{Adam} optimizer with a batch size of $32$, an initial learning rate of $1e-4$ and weight decay of $0.05$.
The learning rate is stepped down by a factor of $0.1$ at the $67\%$ and $89\%$ of training epochs.
We conduct all experiments on $8$ V100 GPUs.

We implemented models with different magnitudes. The settings and performance of these backbones are shown in \Cref{tab:Details}, where $C_i$ denotes the number of channels of the $i$-th DOT stage's output feature, and the number of SA blocks within each DOT stage is also provided. Only one global channel-wise attention block is added to the end of each stage. For each model, accuracy refers to the backbone's accuracy on ImageNet and AP refers to the precision after training on the MS COCO dataset.
All GFLOPs are obtained on the MS COCO dataset.

\subsection{Main Results}

\noindent\textbf{Compare with two-stage/one-stage detection methods.}
The performance of conventional two-stage/one-stage detection methods 
is shown in the first part of \Cref{tab:Comparison}. Overall, anchor-based methods converge fast within only 12 epochs, and the transformer-based methods generally outperform CNN-based methods. For instance, Focal-Tiny-RetinaNet~\cite{FocalTrans} achieves 7.8\% higher AP than the original RetinaNet~\cite{RetinaNet}. However, such good performance comes at the expense of high computational costs; most of these methods need 170 GFLOPs at the minimum. Even the more efficient single-level feature detection method YOLOF~\cite{YOLOF} needs 86 GFLOPs when using ResNet-50 as the backbone.

The performance of our proposed DFFT with 12 epochs is shown in the second part of \Cref{tab:Comparison}. In contrast to the above methods that endure an obvious trade-off between detection precision and inference efficiency, our DFFT can improve these two metrics simultaneously based on its efficient design. For example, \DFFT{NANO} decreases $51\%$ GFLOPs of YOLOF while still increasing $1.4\%$ AP from $37.7\%$ to $39.1\%$, and \DFFT{MEDIUM} achieves $42.6\%$ AP ($4.9\%$ higher than YOLOF) with only $67$ GFLOPs ($22\%$ lower). Furthermore, DFFT reduces $200$ GFLOPs from the best-performed Focal-Tiny-RetinaNet~\cite{FocalTrans} at the cost of merely $1\%$ lower AP.

These comparisons indicate that our DFFT can effectively reduce the computational cost of the inference stage, thereby mitigating the slow inference problem of anchor-based methods without sacrificing the detection precision.


\noindent\textbf{Compare with DETR-based methods.}
The performance of DETR-based methods is shown in the third part of \Cref{tab:Comparison}. We observe that DETR-based methods can achieve better performance and inference efficiency but converges slower. For instance, DETR only needs 86 GFLOPs at the inference stage to achieve 42.0\% AP, but it requires as large as 500 epochs to converge. WB-DETR~\cite{WBDETR} can only achieve 39.6\% AP if given the same training epochs. Subsequent optimized DETR-based methods improve the convergence speed but at the cost of inference efficiency. For example, while Deformable DETR~\cite{DeformableDETR} and Condition DETR~\cite{ConditionalDETR} need 50 epochs to converge and TSP-FOCS~\cite{TSP} and Efficient DETR~\cite{EfficientDETR} need only 36 epochs to converge, their GFLOPs are around $4\%$--$120\%$ larger than DETR.

\begin{figure}[t]
    \centering
    \includegraphics[width=.85\linewidth]{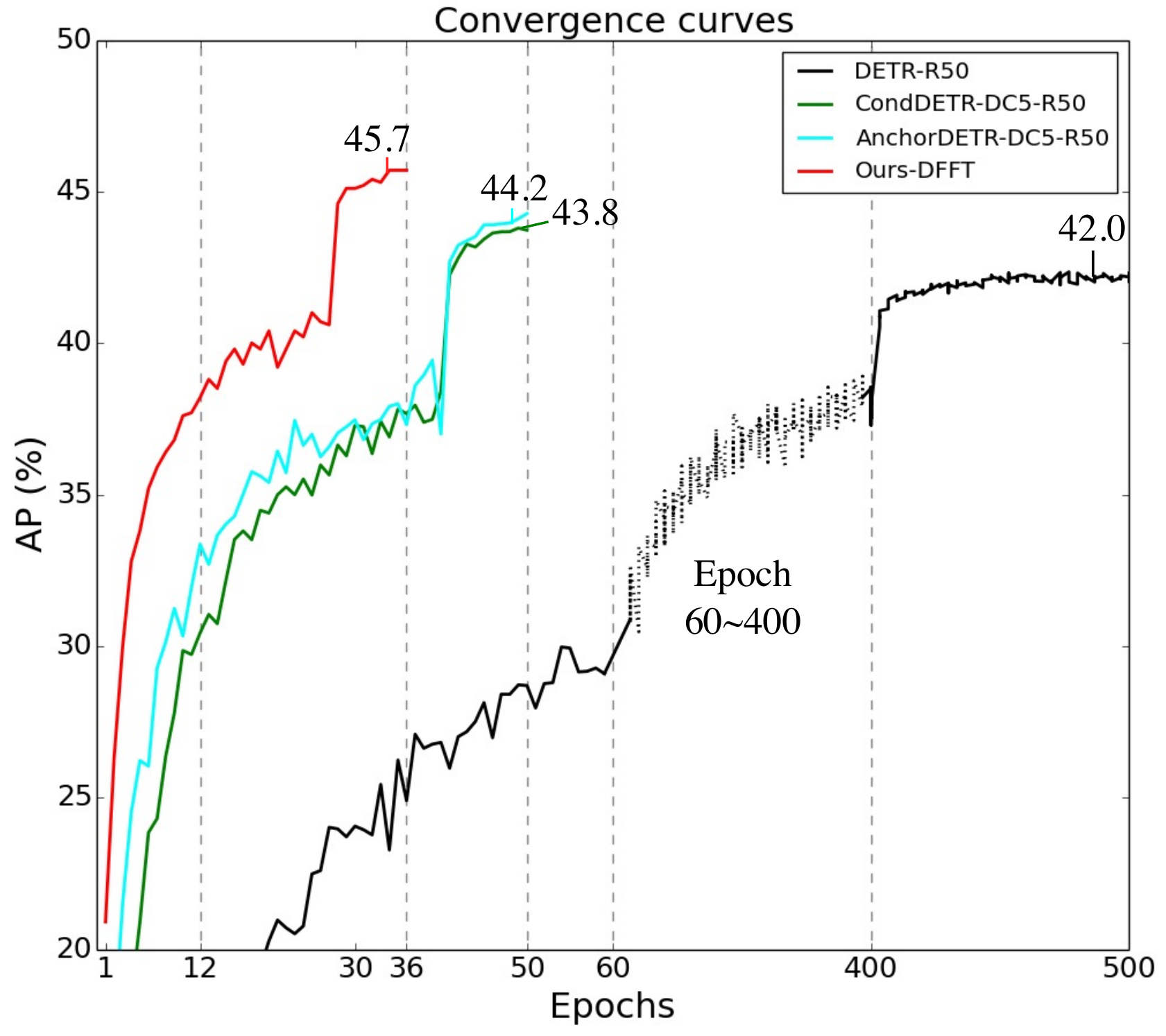}
    \caption{The convergence curves of DFFT and DETR-based methods on the COCO 2017 validation set. Our DFFT converges significantly faster than the cutting-edge DETR-based methods.}
    \label{fig:Convergence}
    \vspace{-1em}
\end{figure}

The performance of our DFFT with 36 epochs is listed in the fourth part of \Cref{tab:Comparison}. In contrast to the DETR-based methods that endure a hard-to-optimize trade-off between the convergence and inference efficiency, our DFFT can achieve state-of-the-art detection prevision without sacrificing neither of these two metrics. Compared with DETR~\cite{DeformableDETR}, our \DFFT{NANO} model improves $13\times$ convergence speed and decreases $51\%$ GFLOPs while achieving significant detection precision ($42.0\%$ vs.\ $42.8\%$). Compared with Efficient DETR~\cite{EfficientDETR} under the same number of training epochs, our model achieves state-of-the-art $45.7\%$ AP with only $67$ GFLOPs ($57\%$ lower).
We further demonstrate the convergence curves of DFFT and DETR-based methods in \Cref{fig:Convergence}. DFFT reduces $28\%$--$92\%$ training epochs of state-of-the-art methods.

These comparisons verify that our DFFT can effectively optimize training and inference efficiency while achieving competitive and even state-of-the-art detection precision.

\begin{figure*}[tbp]
\centering
\begin{minipage}[t]{0.49\linewidth}
	\centering
	\includegraphics[width=0.9\linewidth]{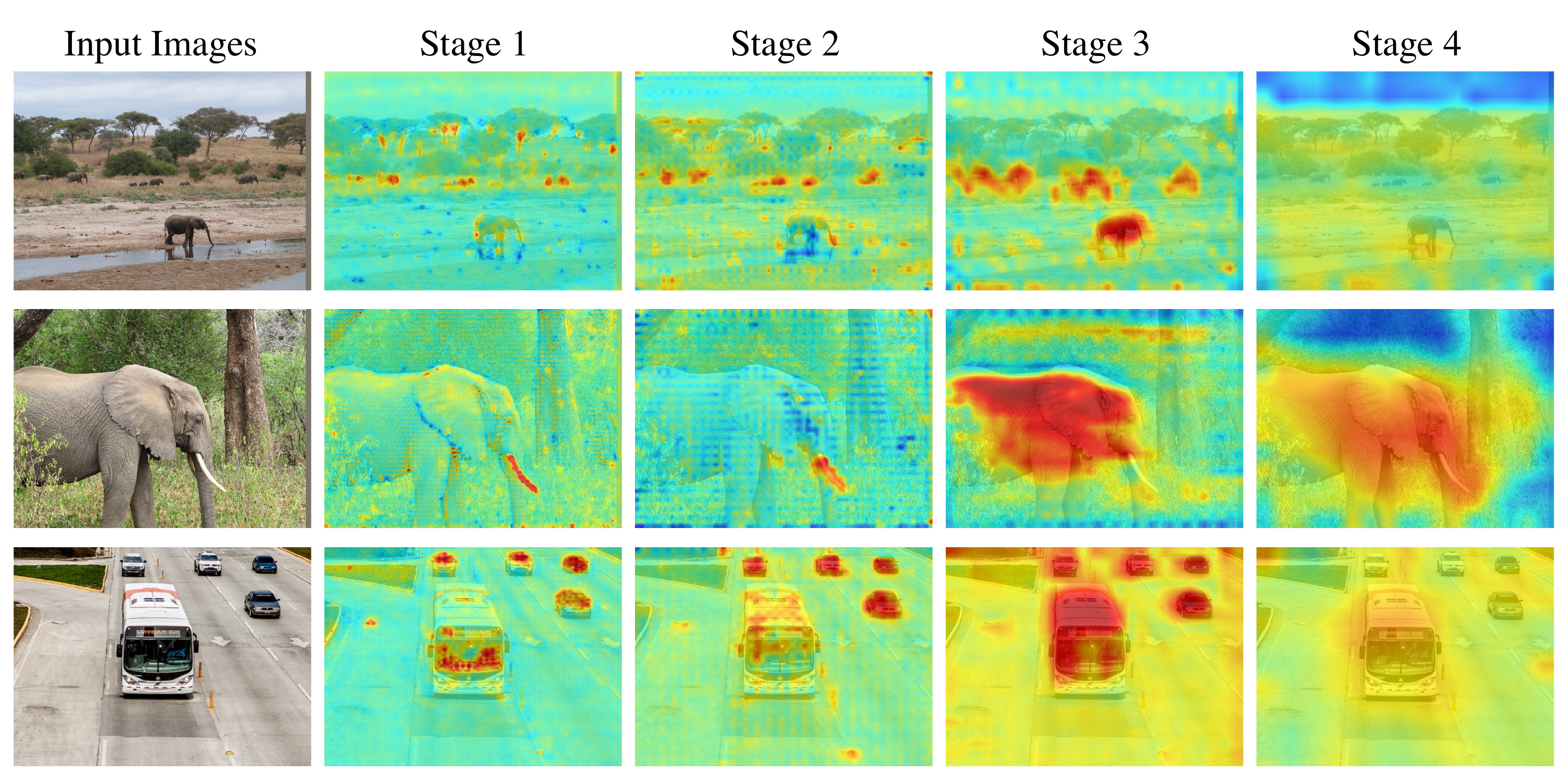}
	\caption{Visualization of the feature map obtained by each DOT backbone stage. The first two stages focus on small objects while the third stage focuses on medium or larger objects. The last stage only responses to large objects.}
	\label{fig:ablationDOT}
\end{minipage}
\hfill
\begin{minipage}[t]{0.49\linewidth}
	\centering
	\includegraphics[width=1.\linewidth]{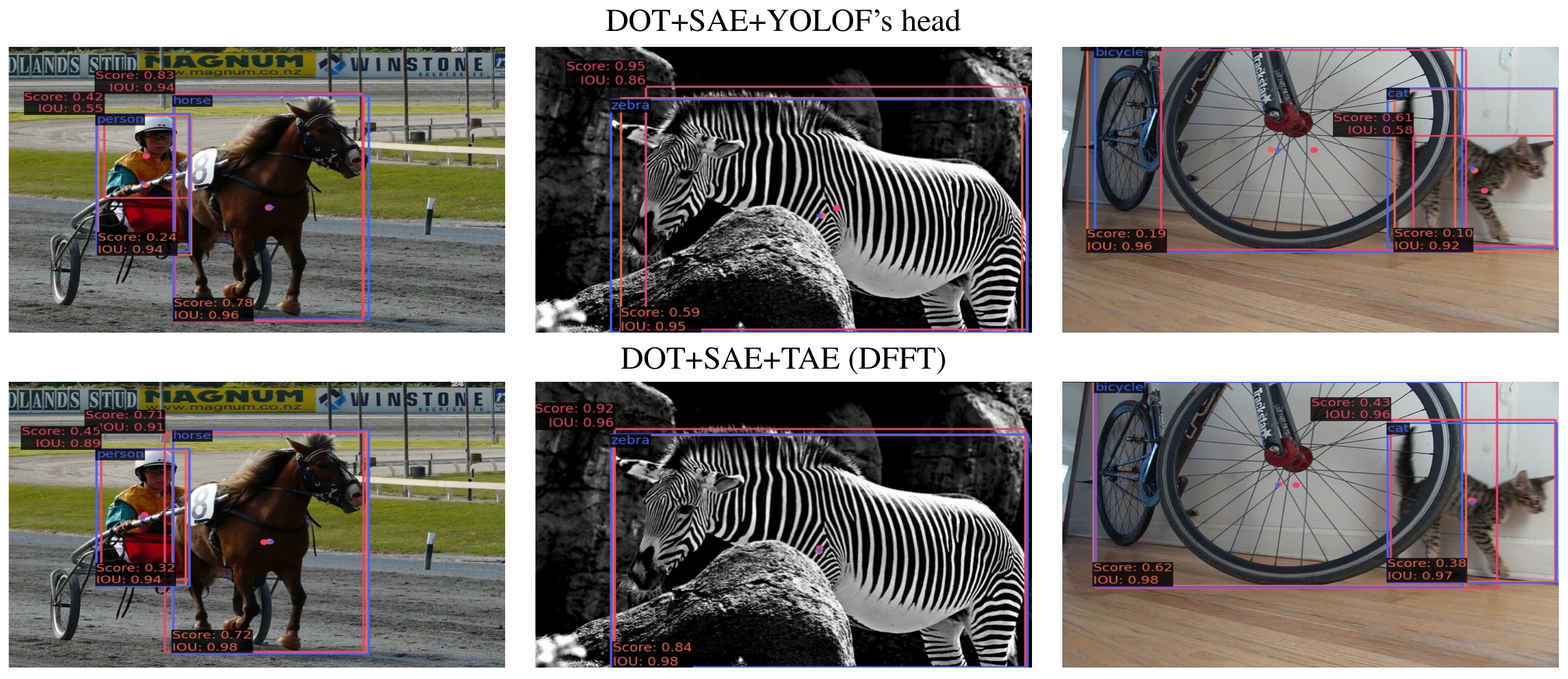}
	\caption{Illustration of detection results from the best anchors for classification (\textcolor{red}{red}) and localization (\textcolor{orange}{orange}). Ground-truth is indicated by \textcolor{blue}{blue} boxes and centers. Our TAE helps provide consistency in predictions of classification and localization.}
	\label{fig:ablationTAE}
\end{minipage}
\vspace{-1em}
\end{figure*}

\subsection{Ablation Study}
We provide a comprehensive ablation study of all designs in our DFFT: the detection-oriented transformer (DOT) backbone, scale-aggregated encoder (SAE), and task-aligned encoder (TAE). The ablation study starts with the major components in DFFT, followed by the specific design of each component. 
We conduct all the experiments on DFFT\textsubscript{SMALL} model that is trained for 12 epochs.

\noindent\textbf{Major Components.}
We first evaluate the efficacy of the major components in DFFT. We disable each component by replacing it with a vanilla method as they are not easily removable from our detection framework. Specifically, we (1) replace the DOT backbone (line 1) with Swin-Transformer~\cite{SwinTransformer} with the similar GFLOPs; (2) disable the SAE module (lines 1, 2, 4) by directly upsampling the last stage's outputs to $\frac{H}{32}\times\frac{W}{32}$ and feed them to TAE; and (3) replace the TAE module (lines 1--3) with YOLOF's head. The results are shown in \Cref{tab:AblationStudy}.

Firstly, the DOT backbone promotes the precision from $33.8\%$ to $37.9\%$, indicating that it can obtain better semantic features that are more suitable for the detection task. Even without the SAE, it gets competitive precisions with only the last stage's outputs. This also suggests that the SAA can capture multi-scale information when aggregating semantic information.

Secondly, SAE further improves the precision to 39.9\% by aggregating multi-scale features into one feature map. Yet, disabling SAE would decrease the precision by 1.6\%.

\begin{table}[tbp]   
    \centering
    \small
    \begin{tabular}{cc|cc}    
        \toprule    
            GCA & SAA &AP (\%) & GFLOPs \\    
        \midrule   
            - & - & 39.0 &48.59 \\
            \YES & -  &40.1 &48.97 \\
            \YES & GCA & 40.4 & 58.65   \\
            \YES & \YES &41.4 &62.23 \\
        \bottomrule   
    \end{tabular}  
    \caption{Ablation study of the DOT backbone. GCA means adding a global channel-wise attention at the end of each DOT block. In the third line, we replace SAA with the same number of GCA.}
    \label{tab:AblationStudyBackbone}
    \vspace{-1em}
\end{table}

Finally, disabling TAE would decrease the prevision by 1.5\%. It verifies the necessity of using TAE to align and encode both the classification and regression features.

\noindent\textbf{Detection-oriented Transformer (DOT) block.}
\Cref{tab:AblationStudyBackbone} studies how the global channel-wise attention (GCA) and semantic-augmented attention (SAA) contributes to DOT's performance. We only modify the backbone network without disabling the SAE and TAE modules.
Lines~1 and~2 show that switching from SW-MSA to our global channel-wise attention can improve $1.1\%$ precision without significant impact on FLOPs.
Lines~2 and~3 show that adding one GCA block costs 11 GFLOPs yet only gains $0.3\%$ AP. Once we replace the GCA block with the SAA module (so that the two settings have the same number of attention nodes), the precision increases from $40.1\%$ to $41.4\%$. These two observations suggest that SAA can enhance performance, and having more attention nodes is not the primary cause.

\begin{table}[tbp]   
    \centering
    \small
    \begin{tabular}{ccc|cc}    
        \toprule   
            DOT  & SAE & TAE & AP (\%) & GFLOPs \\    
        \midrule   
            - & - &- & 33.8 & 45  \\
            \YES & - & - & 37.9 &47  \\
            \YES & \YES & - & 39.9  &58 \\
            \YES & - & \YES & 39.8  &51 \\
            \YES & \YES & \YES  &41.4  &62  \\
        \bottomrule
    \end{tabular}  
    \caption{Ablation study of the three major modules in DFFT.}
    \label{tab:AblationStudy}
    \vspace{-1em}
\end{table}
We further visualize the feature maps from each backbone stage of \DFFT{MEDIUM} trained 12 epochs in \Cref{fig:ablationDOT}. In the first two stages, our DOT backbone can obtain low-level features with sufficient semantic information to capture small objects. The third stage then focuses on medium and large objects. Finally, the last stage only responds to large objects. These observations verify that our DOT backbone can enhance semantic information in low-level features and thus boosting the detection precision.


\begin{table}[t]
\begin{minipage}[t]{.51\linewidth}
  \footnotesize
  \resizebox{.99\linewidth}{!}{
    \begin{tabular}{cc|cc}    
        \toprule    
            SAA & FPN  & AP (\%) & GFLOPs \\    
        \midrule   
            - & - & 37.4 &319 \\
            \YES & - &38.9 & 332 \\
            - & \YES &38.4 & 341  \\ 
        \bottomrule   
    \end{tabular}}
    \caption{\centering Analysis of SAA. }
    \label{tab: Analysis SAA}
\end{minipage}
\begin{minipage}[t]{.47\linewidth}
  \centering
    \footnotesize
    \resizebox{\linewidth}{!}{
    \begin{tabular}{c|cc}    
        \toprule    
            Method & AP (\%) & GFLOPs \\
        \midrule   
            CONCAT &39.6  & 56 \\
            YOLOF &40.3 &58\\
            DFFT & 41.4 &62 \\
        \bottomrule
    \end{tabular}
    }
    \caption{\centering Analysis of SAE}
    \label{tab: Analysis SAE}
\end{minipage}
\vspace{-1em}
\end{table}
\noindent\textbf{Semantic-Augmented Attention (SAA).}
SAA obtains richer low-level semantic features for object detection task by augmenting the semantic information from high-level features to the low-level ones, sharing a similar effect as FPN~\cite{FPN}. For a fair comparison with FPN, we disable the two encoders and directly feed features from the backbone to RetinaNet's head, which is a multi-level feature head that accepts four-scale features. The results are shown in \Cref{tab: Analysis SAA}. While both SAA and FPN improves the precision, SAA obtains 0.5\% higher AP with 9 fewer GFLOPs than FPN. Thus, the global channel-wise attention suits the transformer better than FPN. Including SAA within the forward pass can obtain an even stronger model.

\begin{table}[tb]   
    \centering
    \footnotesize
    \resizebox{0.8\linewidth}{!}{
        \begin{tabular}{l|cc|ccc}
        \toprule
        \multirow{2}{*}{Models} & \multicolumn{2}{c|}{Backbone Settings} &  \multicolumn{3}{c}{}  \\
        & Value of $C_i$ & Number of SA & AP & GFLOPs & FPS \\ \midrule
        DETR-R50 &- &- &42.0 & 86 &24 \\
        Deformable DETR &- &- &43.8 & 173 &14 \\ \midrule
        \DFFT{NANO} & $\p{3, 3, 6, 9}$ & $\p{3, 3, 7, 3}$ &42.8 & 42 & 22 \\
         \DFFT{TINY}&$\p{4, 4, 8, 12}$ & $\p{2, 2, 6, 2}$ &43.5 & 57& 24  \\
         \DFFT{SMALL}& $\p{4, 4, 8, 12}$ & $\p{3, 3, 7, 3}$ &44.5 & 62 & 22  \\
         \DFFT{MEDIUM}& $\p{4, 4, 7, 12}$ & $\p{3, 3, 19, 3}$ &45.7 & 67 & 17  \\
         \DFFT{LARGE}& $\p{6, 6, 8, 12}$ & $\p{3, 3, 19, 3}$ &46.0  &101 & 17  \\ \bottomrule
        \end{tabular}
    }
    \caption{Analysis of how the output feature's number of channels $C_i$ and the number of attention blocks impact GFLOPs and FPS. All the results are measured on the same machine with a V100 GPU using mmdetection~\cite{mmdetection}.}
    \label{tab:AblationFPS}
    \vspace{-1em}
\end{table}
\noindent\textbf{Scale-aggregated Encoder (SAE).}
SAE aggregates multi-scale features into one feature map to reduce the inference stage's computational costs. We compare SAE with a similar design in YOLOF, which exploits a dilated encoder to convert features from multiple scales. \Cref{tab: Analysis SAE} shows that SAE improves $1.1\%$ AP from the dilated encoder of YOLOF. When compared with a vanilla concatenation operation, SAE gets $1.8\%$ higher precision. Overall, SAE can achieve better performance with low computational costs.


\noindent\textbf{Task-aligned Encoder (TAE).}
Benefiting from the group channel-wise attention's capability of modeling semantic relations, TAE handles task conflicts in a coupled head and further generates task-aligned predictions in a single pass. As shown in the first row of \Cref{fig:ablationTAE}, after replacing TAE with YOLOF's head in the baseline model, the best anchors for classification (red) and localization (orange) are distant from each other. That is because
YOLOF~\cite{YOLOF} uses a task-unaligned decoupled head that leads to inconsistent predictions of classification and localization. Comparatively, our TAE provides aligned predictions with both high classification and IOU scores (\emph{e.g.}, person, zebra and cat in \Cref{fig:ablationTAE}).

\noindent\textbf{Analysis of the impact on GFLOPs and FPS.} 
We compare the performance of different models in terms of prediction AP and inference GFLOPs and FPS in \Cref{tab:AblationFPS}. 
For computational performance, GFLOPs and FPS are sensitive to the number of channels and the number of attention blocks, respectively. For example, compared with \DFFT{NANO}, \DFFT{TINY} increases 15 GFLOPs but has a better FPS due to fewer attention blocks, and \DFFT{SMALL} increases 20 GFLOPs but gets a similar FPS. Overall, we achieve better AP and inference efficiency than deformable DETR; at the same FPS, our DFFT has better accuracy and GFLOPS than DETR. Lastly, although we designed the network architecture mainly to reduce GFLOPs, we note that the above observations can also be used to redesign our network and optimize for FPS, which we leave as the future work.






\section{Conclusion}
In this work, we discover a trade-off between training and inference efficiency that hinders transformer-based object detection in large-scale applications. Rather than porting transformers directly to the conventional framework or optimizing the DETR framework, we propose DFFT, a novel design of fully transformer-based object detectors. It enables efficiency in both the training and inference stages for the first time without sacrificing noticeable detection precision. Extensive evaluation reveals our DFFT's unique advantages in capturing low-level semantic features in object detection, as well as its ability to preserve detection precision while trimming the training-inefficient decoders in DETR. Finally, DFFT achieves state-of-the-art performance while using only half the GFLOPs of previous approaches, indicating a promising future work on the large-scale application of transformers in object detection.

{\small
\bibliographystyle{ieee_fullname}
\bibliography{egbib}
}

\end{document}